\documentclass[3p,times,procedia]{elsarticle}
\usepackage{ecrc}

\usepackage{amsmath}
\usepackage{amssymb}
\usepackage{graphicx}
\usepackage{booktabs}
\usepackage[figuresright]{rotating}
\usepackage[bookmarks=false]{hyperref}
    \hypersetup{colorlinks,
      linkcolor=blue,
      citecolor=blue,
      urlcolor=blue}

\usepackage{xurl} 
\volume{00}
\firstpage{1}
\journalname{Transportation Research Procedia}
\runauth{Abdulmumin Sa'ad et al.}
\jid{trpro}

\biboptions{authoryear}

\begin{document}
\begin{frontmatter}

\dochead{}

\title{YOLO-SAT: A Data-based and Model-based Enhanced YOLOv12 Model for Desert Waste Detection and Classification}

\author[a]{Abdulmumin Sa'ad\corref{cor1}}
\author[a]{Sulaimon Oyeniyi Adebayo}

\address[a]{Computer Engineering Department, King Fahd University of Petroleum and Minerals, Dhahran 31261, Saudi Arabia}

\begin{abstract}
The global waste crisis is escalating, with solid waste generation expected to increase tremendously in the coming years.
Traditional waste collection methods, particularly in remote or harsh environments like deserts, are labor-intensive, inefficient, and often hazardous. Recent advances in computer vision and deep learning have opened the door to automated waste detection systems, yet most research focuses on urban environments and recyclable materials, overlooking organic and hazardous waste and underexplored terrains such as deserts. In this work, we propose YOLO-SAT, an enhanced real-time object detection framework based on a pruned, lightweight version of YOLOv12 integrated with Self-Adversarial Training (SAT) and specialized data augmentation strategies. Using the DroneTrashNet dataset, we demonstrate significant improvements in precision, recall, and mean average precision (mAP), while achieving low latency and compact model size suitable for deployment on resource-constrained aerial drones. Benchmarking YOLO-SAT against state-of-the-art lightweight YOLO variants further highlights its optimal balance of accuracy and efficiency. Our results validate the effectiveness of combining data-centric and model-centric enhancements for robust, real-time waste detection in desert environments.
\end{abstract}

\begin{keyword}
Desert waste; Object detection; YOLOv12; Self-Adversarial Training; Data augmentation; DroneTrashNet.
\end{keyword}

\cortext[cor1]{Corresponding author. Tel.: +966-50-857-1045 ; fax: +966-54-090-5091.} 
\end{frontmatter}

\email{g202203620@kfupm.edu.sa (A. Sa'ad), g202203440@kfupm.edu.sa (S. O. Adebayo).}  

\section{Introduction}
Waste generation has been escalating at an alarming rate worldwide, posing significant environmental and societal challenges. Recent studies has shown that in the next 25 years, the amount of waste in less developed nations will increase dramatically \citep{shahriar2024computer}. According to a World Bank study, it was established that without urgent action, global solid waste will increase by roughly 70\% by 2050, reaching 3.4 billion tons annually \citep{abdu2022survey}. Among the categories used to classify waste are plastic, metal, glass, and organic/bio waste. The bulk of household waste is composed of plastic materials, with PET (polyethylene terephthalate), HDPE (high-density polyethylene), and LDPE (low-density polyethylene) accounting for the four main categories \citep{rehman2024framework}. Organic waste constitutes approximately one-third of total waste and can be decomposed into fertile soil. Effective waste disposal involves implementing methods that accelerate the decomposition process. This approach helps minimize the amount of organic waste sent to landfills while simultaneously enhancing recycling rates \citep{shahriar2024computer}.

\begin{figure*}[!t]
\centering
\includegraphics[scale=0.88]{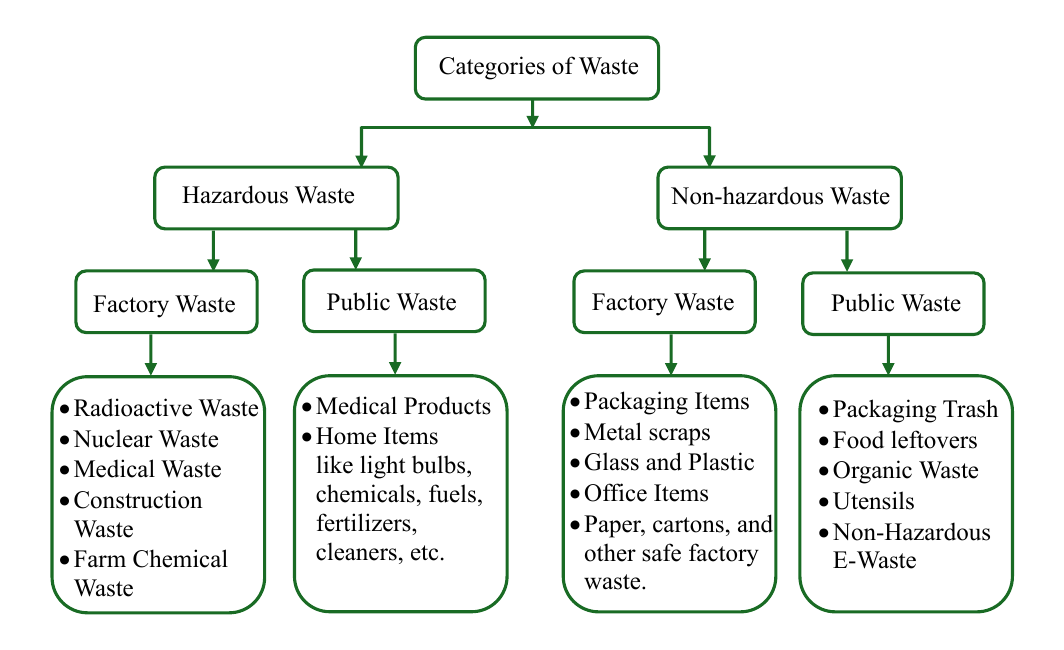}
\caption{Categories of Waste.}
\label{fig:waste_class}
\end{figure*}

Despite increasing public awareness, many individuals still struggle to correctly sort trash into standard recycling categories, such as those defined by researchers like \citep{abdu2022survey} and depicted in Figure~\ref{fig:waste_class}. Waste classification techniques and procedures are applied to important material groups such as paper, glass, metal, wood, and plastic \citep{shahriar2024computer}. The most challenging task, however, is separating different types of materials within a given group. In practice, the burden of sorting and removing litter often falls on municipal workers or volunteers. Traditional waste collection in outdoor environments (including deserts and other remote areas) involves dispatching cleanup teams to manually search for and gather litter. These teams, sometimes called waste pickers, must traverse difficult terrain (on foot or by vehicle) and scan vast areas for trash, a process that is labor-intensive, time-consuming, and potentially dangerous \citep{wang2023deep}. Moreover, such manual efforts are not only inefficient but also expose workers to harsh environmental conditions and hazards. There is a clear need for more efficient, technology-assisted methods to monitor and locate waste in vast areas.

Advances in computer vision and deep learning offer promising solutions to automate waste detection and classification. Image recognition techniques can be employed to identify different types of litter in photographs or video frames, enabling rapid localization of waste without exhaustive human search \citep{majchrowska2022deep, wang2023deep}. Early machine learning approaches to image-based waste recognition relied on hand-crafted features, which often struggled with variability and required significant feature engineering \citep{kumar2024smart, rehman2024framework}. Modern deep learning, by contrast, can automatically learn rich feature representations from data, leading to superior performance in object detection and classification tasks \citep{abdu2022survey,pang2023two,ogrezeanu2024automated}. Convolutional Neural Networks (CNNs) and their variants have revolutionized computer vision by achieving high accuracy in recognizing objects across diverse conditions, all without manual feature design \citep{majchrowska2022deep}. This has catalyzed research into applying deep learning for waste management.

\begin{figure*}[t]
\centering
\includegraphics[scale=0.57]{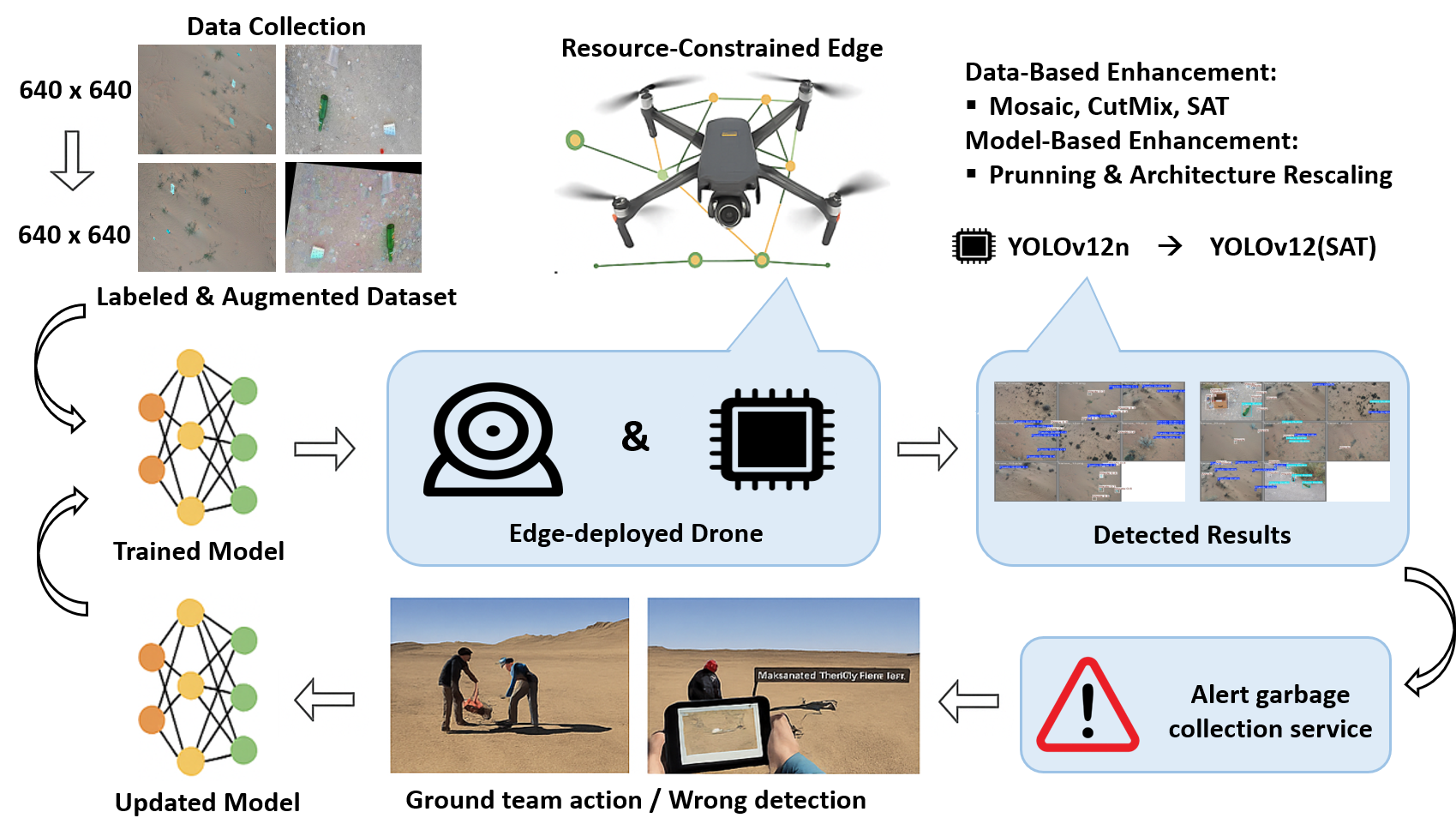}
\caption{Engineering Application Scenario of Desert Waste Detection System.}
\label{fig:scenario}
\end{figure*}

One practical driver of our research is the advent of affordable drones and aerial imaging technology. In recent years, lightweight unmanned aerial vehicles (UAVs) equipped with cameras have become readily available, making it feasible to survey large or hard-to-reach areas from above. Researchers have begun leveraging drones for landscape surveillance to detect litter in environments like coastlines, roadsides, and nature reserves \citep{andriolo2023drones}. Figure~\ref{fig:scenario} demonstrates a scenario and shows how aerial imagery offers a broad vantage point and, when paired with robust object detection algorithms, can quickly pinpoint scattered waste over terrain that would be arduous to patrol on foot. This approach vastly improves the efficiency of identifying cleanup targets and reduces human exposure to hazards. It is especially pertinent to remote and extensive regions such as deserts, where monitoring by humans is difficult.

Desert recreation activities (e.g., off-road safaris and camping) often leave behind litter that not only mars these fragile ecosystems but also contributes to environmental degradation. However, deserts have so far received relatively little attention in the context of automated waste detection, a notable gap this work aims to address. In this paper, we propose an enhanced deep learning model for desert waste detection and classification, building on the state-of-the-art YOLO (You Only Look Once) object detection framework. Specifically,  leverage YOLOv12 to incorporate both data-based and model-based enhancements to improve performance in the challenging desert environment. Data-based enhancements include specialized data augmentation and training strategies to increase the model’s robustness. Model-based enhancements involve architectural and algorithmic modifications to the YOLO detector to better capture varied situations of desert waste.

To address the challenges of detecting camouflaged or partially buried waste in desert environments, we propose a lightweight and efficient object detection pipeline based on a pruned YOLOv12n architecture. We incorporate advanced augmentation techniques (Mosaic \citep{bochkovskiy2020yolov4}, CutMix \citep{yun2019cutmix}), self-adversarial training (SAT) \citep{bochkovskiy2020yolov4}, and “what-not-to-learn” noise injection. These techniques are aimed at improving generalization and reducing false positives, while aggressive pruning ensures the model meets real-time drone deployment constraints. Our enhanced model achieves a significant improvement in detection accuracy (mAP@0.5 up to 94.6\%) while maintaining low latency (16.3 ms) and compact size (4.67 MB).

\section{Related Work}
\textbf{Evolution of the YOLO Object Detection Family}: Real-time object detection has seen tremendous progress over the past decade, led by the evolution of the YOLO family of models \citep{tian2025yolov12}. The original YOLO (v1) \citep{redmon2016you} framed detection as a single-stage regression problem; subsequent versions (v2 \citep{chang2019ship}, v3 \citep{redmon2018yolov3}) added improvements like batch normalization, multi-scale training, and deeper backbones \citep{redmon2017yolo9000}. YOLOv4 \citep{bochkovskiy2020yolov4} introduced a comprehensive “bag of freebies” and “bag of specials,” including CSPDarknet-53, PANet, CIoU loss, CutMix/Mosaic, and Self-Adversarial Training (SAT), achieving superior COCO mAP at real-time speeds. Later versions (v5-v8) focused on engineering optimizations and training pipelines; v7 \citep{wang2023yolov7} added E-ELAN and other tricks, while v8 employed C2f modules and a decoupled head. Emerging work on YOLOv12 \citep{tian2025yolov12} proposes attention-centric designs intended to retain real-time throughput while improving accuracy, especially for small or camouflaged objects \citep{dosovitskiy2020image}.

\textbf{Self-Adversarial Training (SAT)}: introduced with YOLOv4 \citep{bochkovskiy2020yolov4}, is a training-time augmentation where the network first perturbs the input to hide evidence of the object (adversarially), then learns to detect it on the modified image. This improves robustness under occlusion, camouflage, and hard backgrounds and is highly relevant to desert scenes. We adopt SAT to counterbalance performance loss due to pruning and to improve invariance to desert-like artifacts \citep{bochkovskiy2020yolov4}.

\begin{figure*}[t]
\centering
\includegraphics[scale=0.47]{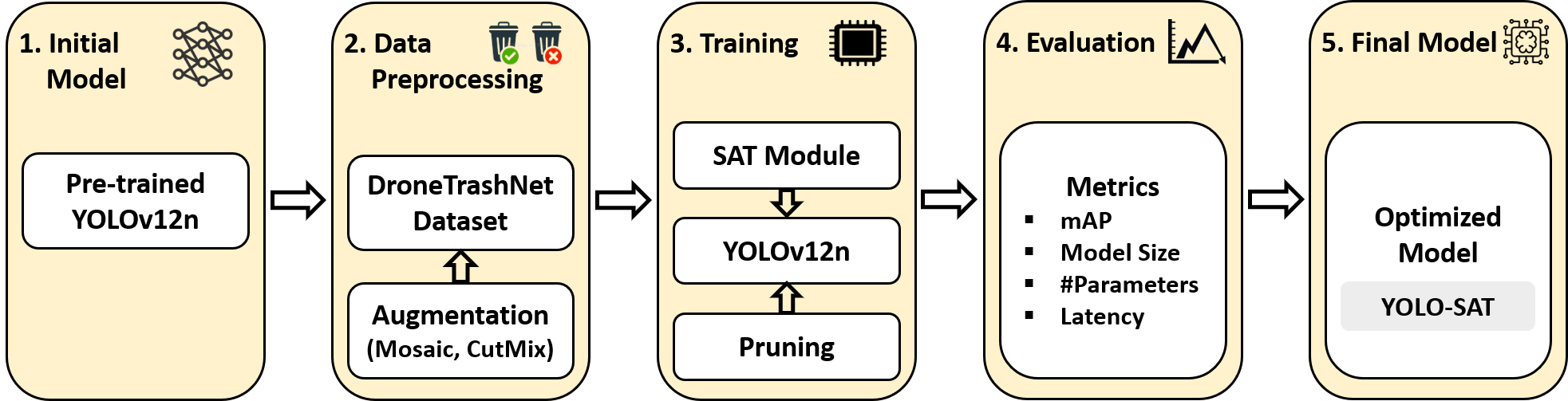}
\caption{Proposed Data-Based and Model-Based Optimization Framework of YOLO-SAT.}
\label{fig:Framework}
\end{figure*}

\textbf{Waste Detection in Desert Environments:}
Most studies focus on urban/controlled environments (e.g., TrashNet \citep{yang2016classification}, TACO \citep{proencca2020taco}). Desert terrains are underexplored and present unique challenges (uniform backgrounds, glare, shadows, partial burial). Prior work includes YOLOv5-based litter detection via UAV imagery \citep{wang2023deep} and the DroneTrashNet dataset/experiments using YOLOv4 with preliminary mAP around 42\% \citep{mehanna2024towards}. These studies motivate domain-specific augmentation, negative examples, and robust training strategies tailored to desert conditions.
\section{Methodology}
Our approach targets a lightweight detector deployable on embedded drone hardware (Figure~\ref{fig:scenario}). We select YOLOv12n and fine-tune on DroneTrashNet \citep{mehanna2024towards} with both data-centric and model-centric strategies: advanced augmentations (Mosaic, CutMix), “what-not-to-learn” noise injection, Self-Adversarial Training, and aggressive pruning (Figure~\ref{fig:Framework}).

\subsection{Experimental Setup}
We combine: (i) noise injection to suppress non-waste patterns \citep{wang2023deep}; (ii) Mosaic/CutMix augmentation \citep{bochkovskiy2020yolov4}; (iii) refined SAT to regain robustness post-pruning; (iv) architectural slimming (channel-width and head pruning); and (v) real-time latency validation.

\subsection{Model Architecture}
\begin{figure}[t]
\centering
\includegraphics[scale=0.9]{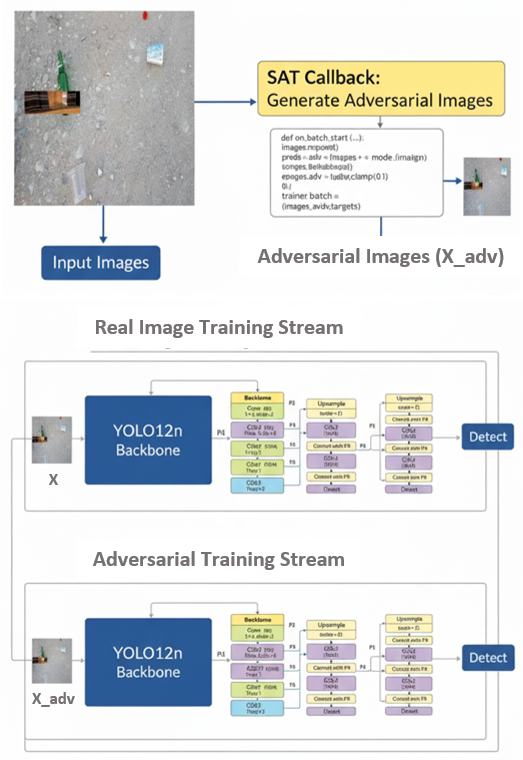}
\caption{Redefined YOLOv12n model architecture with SAT.}
\label{fig:yolo}
\end{figure}

We pruned YOLOv12n (e.g., width multiplier from [0.50, 0.25, 1024] to [0.33, 0.25, 1024]) targeting \(~2.6\,\text{M}\) parameters and \(~6.7\,\text{GFLOPs}\). We incorporate a simple SAT step per batch by adding small image-space perturbations before standard training, which teaches the model to ignore spurious cues and focus on discriminative features. See Figure~\ref{fig:yolo}.

\begin{table}[t]
\caption{Dataset Variants and Augmentation Settings}
\label{tab:dataset_variants}
\begin{tabular*}{\hsize}{@{\extracolsep{\fill}}lccccl@{}}
\toprule
\textbf{Dataset Variant} & \textbf{Total Images} & \textbf{Train} & \textbf{Val} & \textbf{Test} & \textbf{Augmentation Settings} \\
\colrule
Raw     & 200   & 120  & 40  & 40  & None \\
Noisy   & 300   & 180  & 60  & 60  & +100 desert-camp ``negative'' images \\
Aug (mini) & 2,129 & 1,277 & 426 & 426 & NUM\_GEOM = 15, NUM\_CUTMIX = 0, NUM\_MOSAIC = 0 \\
Aug     & 2,129 & 1,277 & 426 & 426 & NUM\_GEOM = 5, NUM\_CUTMIX = 5, NUM\_MOSAIC = 5 \\
\botrule
\end{tabular*}
\end{table}

\subsection{Dataset Generation}
DroneTrashNet \citep{mehanna2024towards} consists of ~200 annotated frames (\textit{plastic bottle}, \textit{glass bottle}, \textit{waste}). We derive four variants (Raw, Noisy, aug (mini), Aug) with a consistent 60/20/20 split as presented in Table~\ref{tab:dataset_variants}, adding desert negatives and advanced augmentations to improve generalization.
\subsection{Impact of Self-Adversarial Training (SAT)}
SAT performs two passes per batch: an adversarial perturbation step on inputs, followed by standard optimization, yielding robustness and compensating for pruning-induced capacity loss \citep{mahto2020refining,chen2020self}. In our context, SAT compensates for accuracy loss from pruning in the lightweight YOLOv12n by reinforcing feature robustness. It continually exposes the model to challenging samples, improving generalization despite reduced parameters.
\subsection{Hardware and Software Setup}
We fine-tune pruned YOLOv12n (n = [0.33, 0.25, 1024]) for 60 epochs (early stop after 15 epochs of no mAP improvement) at 640 resolution and batch size 8, using an NVIDIA RTX A4500 (24 GB VRAM). Tooling: PyTorch/Ultralytics YOLO, NumPy, OpenCV, and Matplotlib.

\section{Results and Discussion}
\subsection{Evaluation and Results}
\begin{figure*}[t]
\centering
\includegraphics[scale=0.55]{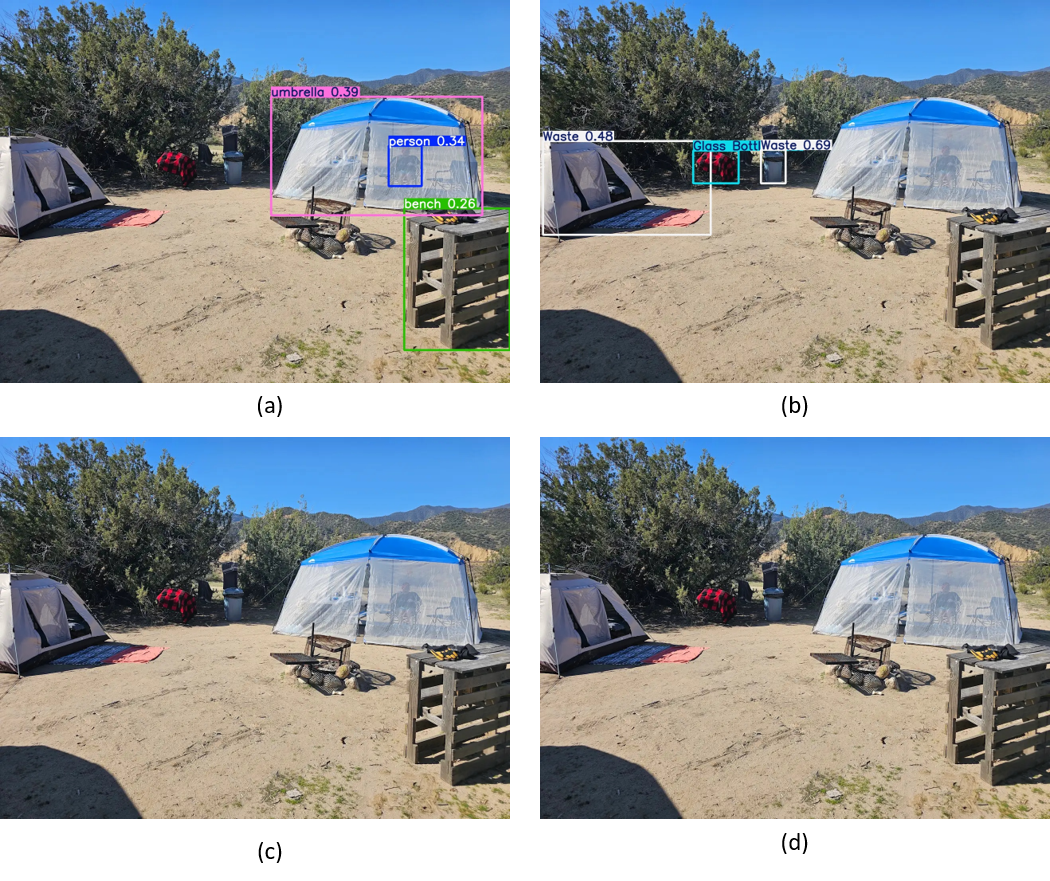}
\caption{Single-class detection across: (a) off-the-shelf YOLOv12n, (b) raw-trained, (c) noisy-trained, (d) augmented-trained.}
\label{fig:single_class}
\end{figure*}

\begin{figure*}[!ht]
\centering
\includegraphics[scale=0.55]{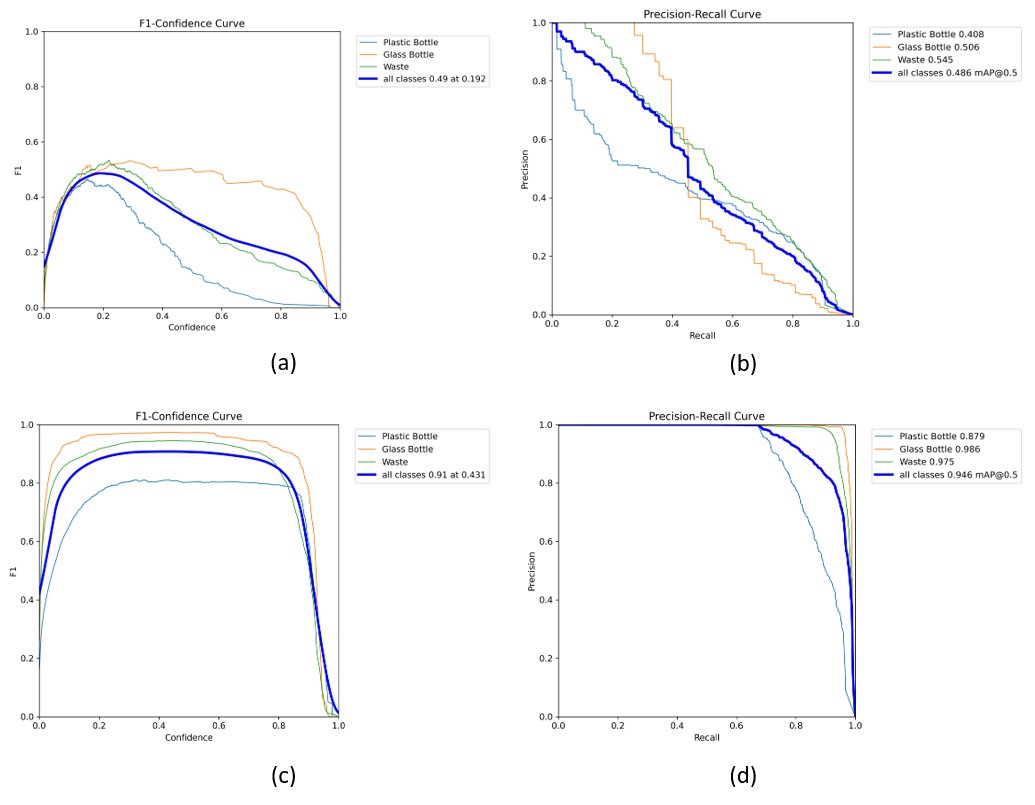}
\caption{F1, precision, and recall trends across dataset variants (a,b) without augmentation (c,d) with augmentation.}
\label{fig:data_based}
\end{figure*}

The evaluation of the proposed YOLO-SAT model for desert waste detection employs standard object detection metrics to ensure accurate performance assessment. The primary metrics include \textit{Precision}, \textit{Recall}, \textit{F1-Score}, \textit{Intersection over Union (IoU)}, and \textit{Mean Average Precision (mAP)}. As presented in Table~\ref{tab:comparison}, these metrics collectively evaluate the model’s balance between accuracy, robustness, and computational efficiency. Figure~\ref{fig:single_class} compares predictions across variants; noise injection reduces false alarms, while full augmentation (Mosaic/CutMix) substantially lifts both precision and recall. As shown in Figure~\ref{fig:data_based}, data augmentation reduced latency from 132.1\,ms to 19.2\,ms and boosted F1 to $\approx$\,0.89 through major precision and recall gains. Incorporating SAT further improved mAP@0.50:0.95 to 0.7783 and lowered latency to 16.3\,ms, while pruning cut parameters and FLOPs to 2.17\,M and 2.91\,GFLOPs. Overall, augmentation and attention jointly enhanced detection accuracy and efficiency.

\begin{table}[t]
\caption{Accuracy, speed, and lightweight characteristics across datasets/models}
\label{tab:comparison}
\begin{tabular*}{\hsize}{@{\extracolsep{\fill}}lllllllll@{}}
\toprule
\textbf{Dataset} & \textbf{Model} & \textbf{mAP@0.50:0.95} & \textbf{mAP@0.50} & \textbf{mAP@0.75} & \textbf{Params} & \textbf{FLOPs} & \textbf{Model Size (MB)} & \textbf{Latency (ms)} \\
\colrule
Raw & YOLOv12n & 0.2719 & 0.4875 & 0.3538 & 2.56\,M & 3.16 & 5.52 & 29.5 \\
Aug (mini) & YOLOv12n & 0.5767 & 0.8583 & 0.6436 & 2.56\,M & 3.16 & 5.52 & 20.3 \\
Aug & YOLOv12n & 0.7850 & 0.9463 & 0.8609 & 2.56\,M & 3.16 & 5.52 & 19.2 \\
Aug & YOLOv12n & 0.7732 & 0.9419 & 0.8523 & 2.17\,M & 2.91 & 4.67 & 17.3 \\
 & (pruned only) &  &  &  &  &  & &  \\
Aug & YOLO-SAT & \textbf{0.7783} & \textbf{0.9409} & \textbf{0.8539} & \textbf{2.17\,M} & \textbf{2.91} & \textbf{4.67} & \textbf{16.3} \\
\botrule
\end{tabular*}
\end{table}

\begin{table}[t]
\caption{Benchmarking of YOLO Lightweight Variants on the Augmented DroneTrashNet Dataset}
\label{tab:benchmarking_results}
\begin{tabular*}{\hsize}{@{\extracolsep{\fill}}lcccccccc@{}}
\toprule
\textbf{Dataset} & \textbf{Model} & \textbf{mAP@0.50:0.95} & \textbf{mAP@0.50} & \textbf{mAP@0.75} & \textbf{Params} & \textbf{FLOPs} & \textbf{Model Size (MB)} & \textbf{Latency (ms)} \\
\colrule
Aug & YOLOv5n       & 0.8030 & 0.9550 & 0.8709 & 2.50\,M & 3.53 & 5.27 & 13.9 \\
Aug & YOLOv8n       & 0.8106 & 0.9541 & 0.8756 & 3.01\,M & 4.04 & 6.25 & 22.7 \\
Aug & YOLOv10n      & 0.7871 & 0.9343 & 0.8472 & 2.27\,M & 3.27 & 5.75 & 26.4 \\
Aug & YOLOv11n      & 0.8057 & 0.9471 & 0.8667 & 2.58\,M & 3.16 & 5.47 & 17.5 \\
Aug & YOLO-SAT & \textbf{0.7783} & \textbf{0.9409} & \textbf{0.8539} & \textbf{2.17\,M} & \textbf{2.91} & \textbf{4.67} & \textbf{16.3} \\
\botrule
\end{tabular*}
\end{table}

\subsection{Benchmarking}
Table~\ref{tab:benchmarking_results} compares lightweight YOLO baselines on the augmented set. YOLOv8n attains the highest mAP but at larger FLOPs/latency; the pruned YOLO-SAT offers the best accuracy--efficiency balance for real-time embedded deployment.

\section{Conclusion}
A carefully balanced combination of data augmentation, noise injection, pruning, and SAT yields a compact YOLOv12n model suitable for robust real-time desert waste detection on drones. High mAP with low latency (\(~16.3\,\text{ms}\)) and small size (\(~4.67\,\text{MB}\)) supports embedded deployment. Future work: expand datasets/classes (including organic/hazardous), temporal tracking, geo-tagging, and further energy-optimized inference.
\section*{Acknowledgments}
The authors acknowledge the support from King Fahd University of Petroleum and Minerals (KFUPM).


\bibliographystyle{elsarticle-harv}
\bibliography{references}

\begin{thebibliography}{23}
\expandafter\ifx\csname natexlab\endcsname\relax\def\natexlab#1{#1}\fi
\providecommand{\url}[1]{\texttt{#1}}
\providecommand{\href}[2]{#2}
\providecommand{\path}[1]{#1}
\providecommand{\DOIprefix}{doi:}
\providecommand{\ArXivprefix}{arXiv:}
\providecommand{\URLprefix}{URL: }
\providecommand{\Pubmedprefix}{pmid:}
\providecommand{\doi}[1]{\href{http://dx.doi.org/#1}{\path{#1}}}
\providecommand{\Pubmed}[1]{\href{pmid:#1}{\path{#1}}}
\providecommand{\bibinfo}[2]{#2}
\ifx\xfnm\relax \def\xfnm[#1]{\unskip,\space#1}\fi
\bibitem[{Abdu and Noor(2022)}]{abdu2022survey}
\bibinfo{author}{Abdu, H.}, \bibinfo{author}{Noor, M.H.M.}, \bibinfo{year}{2022}.
\newblock \bibinfo{title}{A survey on waste detection and classification using deep learning}.
\newblock \bibinfo{journal}{IEEE Access} \bibinfo{volume}{10}, \bibinfo{pages}{128151--128165}.
\bibitem[{Andriolo et~al.(2023)Andriolo, Topouzelis, van Emmerik, Papakonstantinou, Monteiro, Isobe, Hidaka, Kako, Kataoka and Gonçalves}]{andriolo2023drones}
\bibinfo{author}{Andriolo, U.}, \bibinfo{author}{Topouzelis, K.}, \bibinfo{author}{van Emmerik, T.H.M.}, \bibinfo{author}{Papakonstantinou, A.}, \bibinfo{author}{Monteiro, J.G.}, \bibinfo{author}{Isobe, A.}, \bibinfo{author}{Hidaka, M.}, \bibinfo{author}{Kako, S.}, \bibinfo{author}{Kataoka, T.}, \bibinfo{author}{Gonçalves, G.}, \bibinfo{year}{2023}.
\newblock \bibinfo{title}{Drones for litter monitoring on coasts and rivers: suitable flight altitude and image resolution}.
\newblock \bibinfo{journal}{Marine Pollution Bulletin} \bibinfo{volume}{195}, \bibinfo{pages}{115521}.
\newblock \URLprefix \url{https://www.sciencedirect.com/science/article/pii/S0025326X23009554}, \DOIprefix\doi{10.1016/j.marpolbul.2023.115521}.
\bibitem[{Bochkovskiy et~al.(2020)Bochkovskiy, Wang and Liao}]{bochkovskiy2020yolov4}
\bibinfo{author}{Bochkovskiy, A.}, \bibinfo{author}{Wang, C.Y.}, \bibinfo{author}{Liao, H.Y.M.}, \bibinfo{year}{2020}.
\newblock \bibinfo{title}{Yolov4: Optimal speed and accuracy of object detection}.
\newblock \bibinfo{journal}{arXiv preprint arXiv:2004.10934} \URLprefix \url{https://arxiv.org/abs/2004.10934}.
\bibitem[{Chang et~al.(2019)Chang, Anagaw, Chang, Wang, Hsiao and Lee}]{chang2019ship}
\bibinfo{author}{Chang, Y.L.}, \bibinfo{author}{Anagaw, A.}, \bibinfo{author}{Chang, L.}, \bibinfo{author}{Wang, Y.C.}, \bibinfo{author}{Hsiao, C.Y.}, \bibinfo{author}{Lee, W.H.}, \bibinfo{year}{2019}.
\newblock \bibinfo{title}{Ship detection based on yolov2 for sar imagery}.
\newblock \bibinfo{journal}{Remote Sensing} \bibinfo{volume}{11}, \bibinfo{pages}{786}.
\bibitem[{Chen et~al.(2020)Chen, Chen, Zhou, Mao, Li, He, Xue, Zhang and Yu}]{chen2020self}
\bibinfo{author}{Chen, K.}, \bibinfo{author}{Chen, Y.}, \bibinfo{author}{Zhou, H.}, \bibinfo{author}{Mao, X.}, \bibinfo{author}{Li, Y.}, \bibinfo{author}{He, Y.}, \bibinfo{author}{Xue, H.}, \bibinfo{author}{Zhang, W.}, \bibinfo{author}{Yu, N.}, \bibinfo{year}{2020}.
\newblock \bibinfo{title}{Self-supervised adversarial training}, in: \bibinfo{booktitle}{ICASSP 2020-2020 IEEE International Conference on Acoustics, Speech and Signal Processing (ICASSP)}, \bibinfo{organization}{IEEE}. pp. \bibinfo{pages}{2218--2222}.
\bibitem[{Dosovitskiy et~al.(2020)Dosovitskiy, Beyer, Kolesnikov, Weissenborn, Zhai, Unterthiner, Dehghani, Minderer, Heigold, Gelly, Uszkoreit and Houlsby}]{dosovitskiy2020image}
\bibinfo{author}{Dosovitskiy, A.}, \bibinfo{author}{Beyer, L.}, \bibinfo{author}{Kolesnikov, A.}, \bibinfo{author}{Weissenborn, D.}, \bibinfo{author}{Zhai, X.}, \bibinfo{author}{Unterthiner, T.}, \bibinfo{author}{Dehghani, M.}, \bibinfo{author}{Minderer, M.}, \bibinfo{author}{Heigold, G.}, \bibinfo{author}{Gelly, S.}, \bibinfo{author}{Uszkoreit, J.}, \bibinfo{author}{Houlsby, N.}, \bibinfo{year}{2020}.
\newblock \bibinfo{title}{An image is worth 16x16 words: Transformers for image recognition at scale}.
\newblock \bibinfo{journal}{arXiv preprint arXiv:2010.11929} \URLprefix \url{https://arxiv.org/abs/2010.11929}.
\bibitem[{Kumar(2024)}]{kumar2024smart}
\bibinfo{author}{Kumar, D.}, \bibinfo{year}{2024}.
\newblock \bibinfo{title}{Smart garbage detection system for sustainable waste management using deep learning techniques}, in: \bibinfo{booktitle}{2024 IEEE International Conference for Women in Innovation, Technology \& Entrepreneurship (ICWITE)}, \bibinfo{publisher}{IEEE}. pp. \bibinfo{pages}{253--257}.
\bibitem[{Mahto et~al.(2020)Mahto, Garg, Seth and Panda}]{mahto2020refining}
\bibinfo{author}{Mahto, P.}, \bibinfo{author}{Garg, P.}, \bibinfo{author}{Seth, P.}, \bibinfo{author}{Panda, J.}, \bibinfo{year}{2020}.
\newblock \bibinfo{title}{Refining yolov4 for vehicle detection}.
\newblock \bibinfo{journal}{International Journal of Advanced Research in Engineering and Technology (IJARET)} \bibinfo{volume}{11}.
\bibitem[{Majchrowska et~al.(2022)Majchrowska, Mikołajczyk, Ferlin, Klawikowska, Plantykow, Kwasigroch and Majek}]{majchrowska2022deep}
\bibinfo{author}{Majchrowska, S.}, \bibinfo{author}{Mikołajczyk, A.}, \bibinfo{author}{Ferlin, M.}, \bibinfo{author}{Klawikowska, Z.}, \bibinfo{author}{Plantykow, M.A.}, \bibinfo{author}{Kwasigroch, A.}, \bibinfo{author}{Majek, K.}, \bibinfo{year}{2022}.
\newblock \bibinfo{title}{Deep learning-based waste detection in natural and urban environments}.
\newblock \bibinfo{journal}{Waste Management} \bibinfo{volume}{138}, \bibinfo{pages}{274--284}.
\newblock \URLprefix \url{https://www.sciencedirect.com/science/article/pii/S0956053X21006474}, \DOIprefix\doi{10.1016/j.wasman.2021.12.001}.
\bibitem[{Mehanna et~al.(2024)Mehanna, Berchiche, Niare, Maaradji, Hacid and Soukane}]{mehanna2024towards}
\bibinfo{author}{Mehanna, S.}, \bibinfo{author}{Berchiche, N.}, \bibinfo{author}{Niare, A.B.}, \bibinfo{author}{Maaradji, A.}, \bibinfo{author}{Hacid, H.}, \bibinfo{author}{Soukane, A.}, \bibinfo{year}{2024}.
\newblock \bibinfo{title}{Towards real-time image mining for waste detection in deserts}, in: \bibinfo{booktitle}{Innovation and Technological Advances for Sustainability}. \bibinfo{publisher}{CRC Press}, pp. \bibinfo{pages}{352--359}.
\newblock \URLprefix \url{https://www.taylorfrancis.com/chapters/oa-edit/10.1201/9781003496724-34/towards-real-time-image-mining-waste-detection-deserts-souheir-mehanna-nazim-berchiche-assitan-niare-abderrahmane-maaradji-hakim-hacid-assia-soukane}, \DOIprefix\doi{10.1201/9781003496724-34}.
\bibitem[{Ogrezeanu et~al.(2024)Ogrezeanu, Suciu and Itu}]{ogrezeanu2024automated}
\bibinfo{author}{Ogrezeanu, I.A.}, \bibinfo{author}{Suciu, C.}, \bibinfo{author}{Itu, L.M.}, \bibinfo{year}{2024}.
\newblock \bibinfo{title}{Automated waste sorting: A comprehensive approach using deep learning for detection and classification}, in: \bibinfo{booktitle}{2024 32nd Mediterranean Conference on Control and Automation (MED)}, \bibinfo{publisher}{IEEE}. pp. \bibinfo{pages}{268--273}.
\bibitem[{Pang and Huang(2023)}]{pang2023two}
\bibinfo{author}{Pang, H.}, \bibinfo{author}{Huang, C.}, \bibinfo{year}{2023}.
\newblock \bibinfo{title}{A two-stage deep learning framework for enhanced waste detection and classification}, in: \bibinfo{booktitle}{2023 International Conference on Machine Learning and Applications (ICMLA)}, \bibinfo{publisher}{IEEE}. pp. \bibinfo{pages}{2014--2021}.
\bibitem[{Proen{\c{c}}a and Sim{\~o}es(2020)}]{proencca2020taco}
\bibinfo{author}{Proen{\c{c}}a, P.F.}, \bibinfo{author}{Sim{\~o}es, P.}, \bibinfo{year}{2020}.
\newblock \bibinfo{title}{Taco: Trash annotations in context for litter detection}.
\newblock \bibinfo{journal}{arXiv preprint arXiv:2003.06975} .
\bibitem[{Redmon et~al.(2016)Redmon, Divvala, Girshick and Farhadi}]{redmon2016you}
\bibinfo{author}{Redmon, J.}, \bibinfo{author}{Divvala, S.}, \bibinfo{author}{Girshick, R.}, \bibinfo{author}{Farhadi, A.}, \bibinfo{year}{2016}.
\newblock \bibinfo{title}{You only look once: Unified, real-time object detection}, in: \bibinfo{booktitle}{Proceedings of the IEEE Conference on Computer Vision and Pattern Recognition (CVPR)}, \bibinfo{publisher}{IEEE}. pp. \bibinfo{pages}{779--788}.
\newblock \URLprefix \url{https://www.cv-foundation.org/openaccess/content_cvpr_2016/html/Redmon_You_Only_CVPR_2016_paper.html}.
\bibitem[{Redmon and Farhadi(2017)}]{redmon2017yolo9000}
\bibinfo{author}{Redmon, J.}, \bibinfo{author}{Farhadi, A.}, \bibinfo{year}{2017}.
\newblock \bibinfo{title}{Yolo9000: Better, faster, stronger}, in: \bibinfo{booktitle}{Proceedings of the IEEE Conference on Computer Vision and Pattern Recognition (CVPR)}, \bibinfo{publisher}{IEEE}. pp. \bibinfo{pages}{7263--7271}.
\newblock \URLprefix \url{https://openaccess.thecvf.com/content_cvpr_2017/html/Redmon_YOLO9000_Better_Faster_CVPR_2017_paper.html}.
\bibitem[{Redmon and Farhadi(2018)}]{redmon2018yolov3}
\bibinfo{author}{Redmon, J.}, \bibinfo{author}{Farhadi, A.}, \bibinfo{year}{2018}.
\newblock \bibinfo{title}{Yolov3: An incremental improvement}.
\newblock \bibinfo{journal}{arXiv preprint arXiv:1804.02767} \URLprefix \url{https://arxiv.org/abs/1804.02767}.
\bibitem[{Rehman and Deriche(2024)}]{rehman2024framework}
\bibinfo{author}{Rehman, A.}, \bibinfo{author}{Deriche, M.}, \bibinfo{year}{2024}.
\newblock \bibinfo{title}{A framework for segmenting and classification of plastic waste using deep networks}, in: \bibinfo{booktitle}{2024 21st International Multi-Conference on Systems, Signals \& Devices (SSD)}, \bibinfo{publisher}{IEEE}. pp. \bibinfo{pages}{349--353}.
\bibitem[{Shahriar et~al.(2024)Shahriar, Shammi, Ahmed and Rahaman}]{shahriar2024computer}
\bibinfo{author}{Shahriar, M.F.}, \bibinfo{author}{Shammi, S.J.}, \bibinfo{author}{Ahmed, M.S.}, \bibinfo{author}{Rahaman, M.A.}, \bibinfo{year}{2024}.
\newblock \bibinfo{title}{Computer vision based waste classification and management system}, in: \bibinfo{booktitle}{2024 IEEE International Conference on Computing, Applications and Systems (COMPAS)}, \bibinfo{publisher}{IEEE}. pp. \bibinfo{pages}{1--6}.
\bibitem[{Tian et~al.(2025)Tian, Ye and Doermann}]{tian2025yolov12}
\bibinfo{author}{Tian, Y.}, \bibinfo{author}{Ye, Q.}, \bibinfo{author}{Doermann, D.}, \bibinfo{year}{2025}.
\newblock \bibinfo{title}{Yolov12: Attention-centric real-time object detectors}.
\newblock \bibinfo{journal}{arXiv preprint arXiv:2502.12524} \URLprefix \url{https://arxiv.org/abs/2502.12524}.
\bibitem[{Wang et~al.(2023a)Wang, Bochkovskiy and Liao}]{wang2023yolov7}
\bibinfo{author}{Wang, C.Y.}, \bibinfo{author}{Bochkovskiy, A.}, \bibinfo{author}{Liao, H.Y.M.}, \bibinfo{year}{2023}a.
\newblock \bibinfo{title}{Yolov7: Trainable bag-of-freebies sets new state-of-the-art for real-time object detectors}, in: \bibinfo{booktitle}{Proceedings of the IEEE/CVF Conference on Computer Vision and Pattern Recognition (CVPR)}, \bibinfo{publisher}{IEEE}. pp. \bibinfo{pages}{7464--7475}.
\newblock \URLprefix \url{https://openaccess.thecvf.com/content/CVPR2023/html/Wang_YOLOv7_Trainable_Bag-of-Freebies_Sets_New_State-of-the-Art_for_Real-Time_Object_Detectors_CVPR_2023_paper.html}.
\bibitem[{Wang et~al.(2023b)Wang, Leonce, Hacid and Edirisinghe}]{wang2023deep}
\bibinfo{author}{Wang, G.}, \bibinfo{author}{Leonce, A.}, \bibinfo{author}{Hacid, H.}, \bibinfo{author}{Edirisinghe, E.A.}, \bibinfo{year}{2023}b.
\newblock \bibinfo{title}{Deep neural network based automatic litter detection in desert areas using unmanned aerial vehicle imagery}, in: \bibinfo{booktitle}{2023 International Symposium on Networks, Computers and Communications (ISNCC)}, \bibinfo{publisher}{IEEE}. pp. \bibinfo{pages}{1--8}.
\bibitem[{Yang and Thung(2016)}]{yang2016classification}
\bibinfo{author}{Yang, M.}, \bibinfo{author}{Thung, G.}, \bibinfo{year}{2016}.
\newblock \bibinfo{title}{Classification of trash for recyclability status}.
\newblock \bibinfo{journal}{CS229 project report} \bibinfo{volume}{2016}, \bibinfo{pages}{3}.
\bibitem[{Yun et~al.(2019)Yun, Han, Oh, Chun, Choe and Yoo}]{yun2019cutmix}
\bibinfo{author}{Yun, S.}, \bibinfo{author}{Han, D.}, \bibinfo{author}{Oh, S.J.}, \bibinfo{author}{Chun, S.}, \bibinfo{author}{Choe, J.}, \bibinfo{author}{Yoo, Y.}, \bibinfo{year}{2019}.
\newblock \bibinfo{title}{Cutmix: Regularization strategy to train strong classifiers with localizable features}, in: \bibinfo{booktitle}{Proceedings of the IEEE/CVF international conference on computer vision}, pp. \bibinfo{pages}{6023--6032}.

\end{thebibliography}

\end{document}